# Design and Validation of a Low-Cost Smartphone Based Fluorescence Detection Platform Compared with Conventional Microplate Readers

**Zhendong Cao**
EIM Technology
zcao@eimtechnology.com

**Katrina G. Salvante**
Faculty of Health Sciences,
Simon Fraser University
kgsalvan@sfu.ca

**Ash Parameswaran**
School of Engineering Science,
Simon Fraser University
paramesw@sfu.ca

**Pablo A. Nepomnaschy**
Faculty of Health Sciences
Simon Fraser University
pablo_nepomnaschy@sfu.ca

**Hongji Dai**
EIM Technology
tdai@eimtechnology.com

*Abstract* – A low cost fluorescence-based optical system is developed for detecting the presence of certain microorganisms and molecules within a diluted sample. A specifically designed device setup compatible with conventional 96 well plates is chosen to create an ideal environment in which a smart phone camera can be used as the optical detector. In comparison with conventional microplate reading machines such as Perkin Elmer Victor Machine, the device presented in this paper is not equipped with expensive elements such as exciter filer, barrier filter and photomultiplier; instead, a phone camera is all needed to detect fluorescence within the sample. The strategy being involved is to determine the relationship between the image color of the sample in RGB color space and the molar concentration of the fluorescence specimen in that sample. This manuscript is a preprint version of work related to a publication in IEEE. The final version may differ from this manuscript.

## I. Introduction

Fluorescence-based imaging techniques have been widely applied in biological imaging. When an fluorescent object absorbed excitation energy, the emitted photons have longer wavelength than the excitation photons; the difference of the wavelengths at maximum spectral intensities between emission spectrum and excitation spectrum is known as the stoke shift. The stoke shift which usually ranges from 10nm to 50nm allows the fluorescent object to be distinguished and detected using imaging techniques [1]. Conventional fluorescence microscopes use optical filter to limit the emission detection band so that the detector will only pick up photons within the correct range [2]. The fact that lower molar concentration of fluorophore leads to lower fluorescent emission intensity makes detection of emitted photons being difficult [3]. For commercial instrument, a photomultiplier is usually added to dramatically boost up the photon sensitivity of the detector [4].

Instead of equipping an expensive photomultiplier, a different strategy based on RGB color system is used in this paper. According to the CIE 1931 chromaticity diagram in Figure 1, the changing in photon wavelengths will result in changing in color. Theoretically, if the excitation light remains the same and a noise free environment is provided, it is possible to determine the changing of fluorescence intensity in response to the fluorophore molarity variations using the color system only.

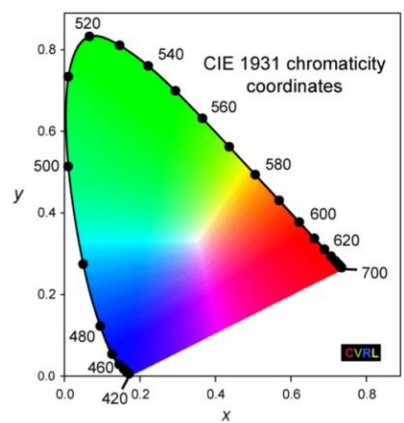

Figure 1. The CIE 1931 colour space chromaticity diagram (public domain).

In this paper we built a device which uses a phone camera as the fluorescence detector. Most of the phone camera nowadays such as iPhone and Samsung Galaxy have 24-bit color depth therefore 8-bit is assigned for each of the R, G, B channel [5]. With 8-bit color depth per channel, the camera is able to distinguish over 16 million colors. Though a digital camera is excellent in color representation, it is not able to distinguish between green fluorescent photons and a green color since they all appear "fluorescent" in terms of color. To avoid such situation happens, all materials or objects that might create "fake fluorescent colors" must either be eliminated from the setup or specially processed so that it contains no overlapping color elements with the color of the emission light source. To validate the performance of our device, we conducted experiments on different fluorescence samples and compared results with commercial microplate reader in the following sections.

## II. Experimental Setup

### A. Test Device Setup

The device with each component being displayed is shown in Figure 2. The LED array and 96 well plate will be placed inside the plate housing. The excitation UV light source can hit the fluorescent specimen inside the 96 well plate in an orthogonal direction; in this manner, the scattered light of the fluorescent solution is minimized [6]. The UV LED has wavelength around 400nm which is enough to excite fluorescein and green fluorescent protein [7], [8].

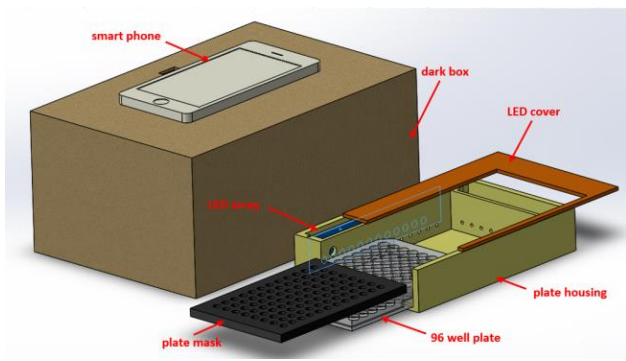

Figure 2. Device Setup with all components being displayed

While conducting the experiment, we applied the LED cover, plate mask to reduce the scattering excitation light noise and side wall color noise (explained in Section III). The plate mask is nicely painted to black color so that it contains no green element. Finally, the entire setup will be covered by the dark box to prevent light pollution from outside. The small window on top of the dark box is for photo acquisition from a phone camera.

### B. Fluorescence Specimen Setup

a) Fluorescein

The fluorescein sample with 100mM molar concentration was made by dissolving 37.6mg of fluorescein sodium salt into 1000uL of water [9]. A fold-10 dilution was applied so that the fluorescein molarity corresponding to each well is listed as Figure3:

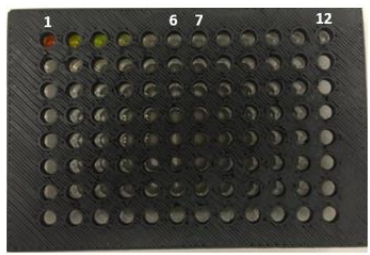

Figure 3. Fluorescein sample preparing in the 96 well plate. Left: the fluorescein molarity corresponding to each well. Right: the 96 well plate covered by the plate mask

The mapping between well position index and fluorescein concentrations in Figure 3 will be referred in section IV. Due to the fact that fluorescein molarity is consistently decreasing, we expect the measurement of fluorescein intensity decreases from index 1 to index 12.

b) Green Fluorescent Protein (GFP)

We use GFP yeast as the test sample. The test was split into two groups where each group contains different concentrations of GFP yeast and yeast control:

Group 1:

GFP yeast with higher concentration, m

Yeast control with higher concentration, mc

Group 2:

GFP yeast with lower concentration, n

Yeast control with lower concertation, nc

In this stage, the values for m, n, mc and nc are not being quantified yet, but their relations are as stated:

$$m \approx 3 * n$$

$$mc \approx 3 * nc$$

We prepared two clean 96 well plates and carefully transferred samples from group 1 and group 2 in each plate. A fold-10 dilution was made for each group; the concentrations of solution for each well are mapped as shown in Table I:

Table I. Yeast samples in Group 1 and Group 2

| Well Index | Group 1 Sample | | Group 2 Sample | |
|---|---|---|---|---|
| | GFP High | Control High | GFP Low | Control Low |
| 1 | m | | n | |
| 2 | m/10 | | n/10 | |
| 3 | m/100 | | n/100 | |
| 4 | m/1000 | | n/1000 | |
| 5 | m/10000 | | n/10000 | |
| 6 | | mc | | nc |
| 7 | water | | water | |

The diluted GFP yeast are placed in the 1st to 5th wells of each plate accordingly; the 6th well has the control yeast and the 7th well has water. Once the dilutions are prepared, we tested two groups of samples in both our proposed device and Victor machine.

## III. Experimental Methods

### A. Image Acquisition

The camera shall be positioned at the right top of each well while taking pictures. Wells that are not axially aligned with camera may contribute side wall noise as indicated in Figure 4 (a) and (b).

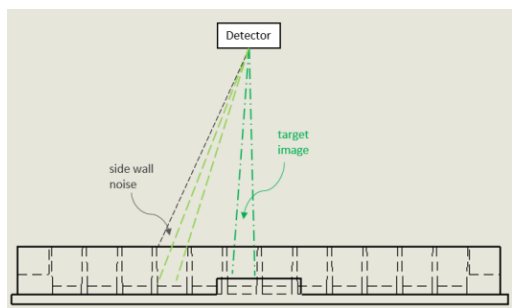

(a) Geometric explanation of side wall noise

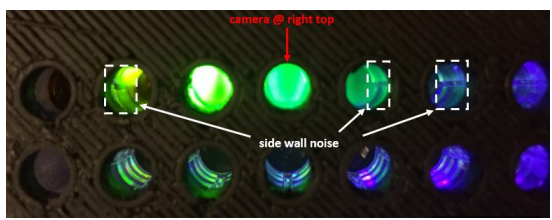

(b) Actual effects of side wall noise

Figure 4. Side wall noise arise from the view angle of camera

The experiment presented in this paper is only for a single row (12 wells) test in a 96 well plate. Hence we need at most 12 images for one test.

B. Image Signal Processing

We developed a Matlab algorithm which intakes camera images and then generates RGB pixel value profiles for further analysis. Each collected sample image will be processed by the algorithm explained in chart 1.

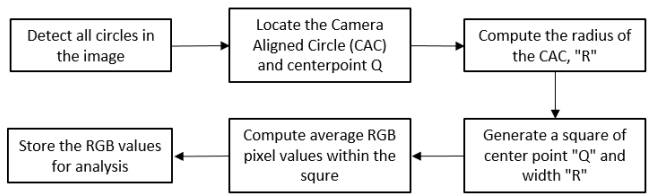

Chart 1. Matlab algorithm to extract image segments in the target well

IV. Experimental Results

A. Fluorescein Results

The diluted fluorescein samples are measured using the Victor Machine and our device; the result data is plotted in logarithm scales in Figure 5 (a) and (b).

Data of the well index 1 and 2 can be ignored since their molar concentrations exceed the measuring range. From well 3 to well 8, the declining tendency of fluorescein intensity measured by Victor Machine can be easily observed from graph; the values corresponding to well [9,12] are [1364, 1205, 1028, 1083] respectively. The measuring result holds valid until well 10; the intensity in well 11 is not reliable since it is lower than water. Based on this test result, we claim that the maximum fluorescence detection range of Victor Machine is 100pM.

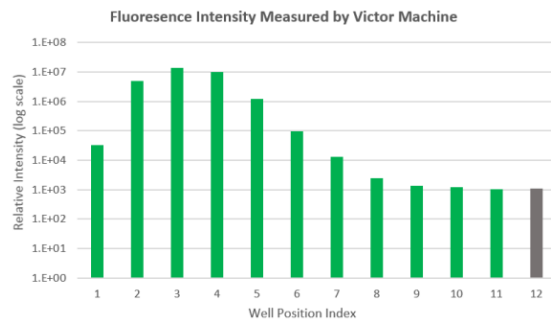

(a) From Victor Machine

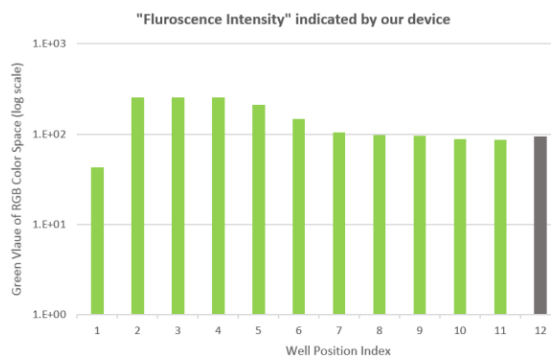

(b) From our device

Figure 5. Fluorescein test results obtained from both the Victor Machine and our device, the 12$^{th}$ index is colored in gray to present water whereas others having fluorescein samples

From the test result using our device, we also observed a consistent decline tendency of green values of the RGB color space) in response to the decreasing in fluorescein molarity. In Figure 5 (b), the green value at well 7 (100nM) is higher than 100 whereas the rest (from well 8 to well 12) are lower than 100. The values in well [8, 12] are measured as [98.45, 96.89, 88.95, 87.31, 93.85], though the green value at well 8 can be distinguished from others by at least 1.6%, this tiny difference is not evident enough by considering the errors occurred in our test setup and measuring approaches. In this stage, we claim the detection range of our device is 100nM.

B. GFP Yeast Results

Figure 6 (a) and (b) has the result obtained from Victor Machine and our device respectively. Referring to Table I, the first 5 wells contain the fold-10 diluted GFP yeast samples. The yeast control is injected into well 6 to serve as a reference guide. We also included water in the test for comparison (well 7).

Since non-GFP yeast control does not glow, the reading of its fluorescence intensity shall be lower than any other GFP yeast samples. Based on this fact, Victor Machine interprets correct intensity readings up to well 4 (n/1000) for Group 1 samples and up to well 3 (n/100) for Group 2 samples. By considering the overall performance, Victor Machine can measure GFP yeast up to (n/100) %.

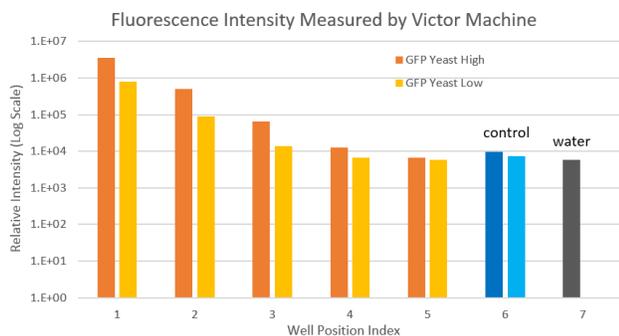

(a) Victor Machine Result

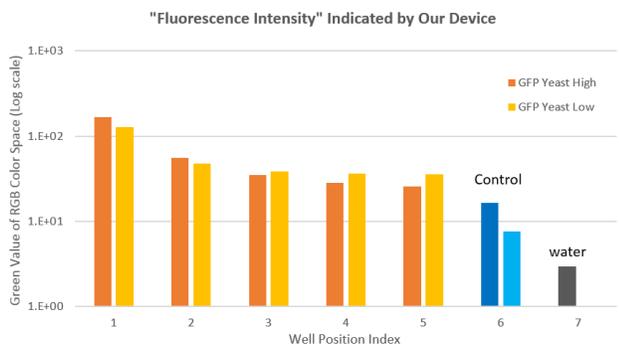

(b) Our Device Result

Figure 6. GFP Yeast test results obtained using Victor Machine and our device, the 6th well contains yeast control and the 7th well has water

On the other hand, the test results obtained by our device are correctly showing that the green values of all GFP yeast samples are higher than yeast control. However, in well 3, the graph indicates that the Group 1 sample has lower intensity than Group 2 sample; this indication is wrong since we know m% > n%. Therefore, the result for our device is only valid up to well 2, or (n/10) %.

The experimental results on fluorescein and GFP yeast test are summarized in Table II,

Table II. Summary of the fluorescein and GFP yeast experimental results

| Sample | Victor Machine | Our Device |
|---|---|---|
| Fluorescein | 100 pM | 100 nM |
| GFP Yeast | n/100 M | n/10 M |

## V. Conclusion and Future Work

A low cost, portable and compact fluorescence-based optical system was introduced in this paper. Based on the fluorescein and GFP yeast experimental results, this system demonstrated the capability to detect the presence of fluorescence objects in a solution sample. We tested the measuring accuracy of our device by comparing the experimental results with a conventional microplate reader. Figure 5 and Figure 6 show that the green channel values exhibited in our experimental results match to the fluorescence intensity measured by the Victor Machine. Though our device has a lower detection range than the Victor Machine by a factor of 1000 in fluorescein test and 10 in GFP yeast test (as shown in Table II), it is significantly inexpensive, portable and accessible. The device is still in the early stage of the design, improvements on the hardware setup and signal processing algorithm will be made in next stage to increase the signal to noise ratio of the image. More experiments on different fluorescent materials will be conducted to give a quantitative relationship between the fluorescence intensity and molar concentration of the fluorescent sample.

### Acknowledgement

The authors would like to thank Dr. Katrina G. Salvante from the Faculty of Health Sciences, SFU, for providing us with the laboratory guide and assistance for preparing fluorescein and GFP yeast samples.